\documentclass[sigconf]{acmart}
\AtBeginDocument{%
  }

\copyrightyear{2025} \acmYear{2025} \setcopyright{rightsretained} \acmConference[SIGIR '25]{Proceedings of the 48th International ACM SIGIR Conference on Research and Development in Information Retrieval}{July 13--18, 2025}{Padua, Italy} \acmBooktitle{Proceedings of the 48th International ACM SIGIR Conference on Research and Development in Information Retrieval (SIGIR '25), July 13--18,2025, Padua, Italy}\acmDOI{10.1145/3726302.3730143} \acmISBN{979-8-4007-1592-1/2025/07}

\begin{document}

\title {CoachGPT: A Scaffolding-based Academic Writing Assistant}

\author{Fumian Chen}
\email{fmchen@udel.edu}
\affiliation{%
  \institution{Institute for Financial Services Analytics \\ University of Delaware}
  \city{Newark}
  \state{Delaware}
  \country{USA}
}

\author{Sotheara Veng}
\email{sotheara@udel.edu}
\affiliation{%
  \institution{School of Education \\ University of Delaware}
  \city{Newark}
  \state{Delaware}
  \country{USA}
}

\author{Joshua Wilson}
\email{joshwils@udel.edu}
\affiliation{%
  \institution{School of Education \\ University of Delaware}
  \city{Newark}
  \state{Delaware}
  \country{USA}
}

\author{Xiaoming Li}
\email{xli@udel.edu}
\affiliation{%
  \institution{Department of Electrical and Computer Engineering \\ University of Delaware}
  \city{Newark}
  \state{Delaware}
  \country{USA}
}

\author{Hui Fang}
\email{hfang@udel.edu}
\affiliation{%
  \institution{Department of Electrical and Computer Engineering \\ Institute for Financial Services Analytics\\ University of Delaware}
  \city{Newark}
  \state{Delaware}
  \country{USA}
}

\renewcommand{\shortauthors}{Fumian Chen, Sotheara Veng, Joshua Wilson, Xiaoming Li, and Hui Fang}

\begin{abstract}
  Academic writing skills are crucial for students' success but can feel overwhelming without proper guidance and practice, particularly when writing in a second language. Traditionally, students ask instructors or search dictionaries, which are not universally accessible. Early writing assistants emerged as rule‑based systems that focused on detecting misspellings, subject‑verb disagreements, and basic punctuation errors but are inaccurate and lack contextual understanding. Machine learning-based assistants demonstrate a strong ability for language understanding but are expensive to train. Large language models (LLMs) have shown remarkable capabilities in generating responses in natural languages based on given prompts, but they have a fundamental limitation in education: they generate essays without teaching, which can have detrimental effects on learning when misused. To address this limitation, we develop {\em CoachGPT}, which leverages LLMs to assist academic writing for those with limited educational resources and those who prefer self-paced learning. CoachGPT is an AI agent-based web application that (1) takes instructions from experienced educators, (2) converts instructions into sub-tasks, and (3) provides real-time feedback and suggestions using large language models. This unique scaffolding structure makes CoachGPT unique among existing writing assistants. Compared with existing writing assistants, CoachGPT provides a more immersed writing experience with personalized messages. Our user studies prove the usefulness of CoachGPT and the potential of large language models for academic writing. 
\end{abstract}




\maketitle

\begin{figure*}
    \centering
    \includegraphics[width=0.98\linewidth]{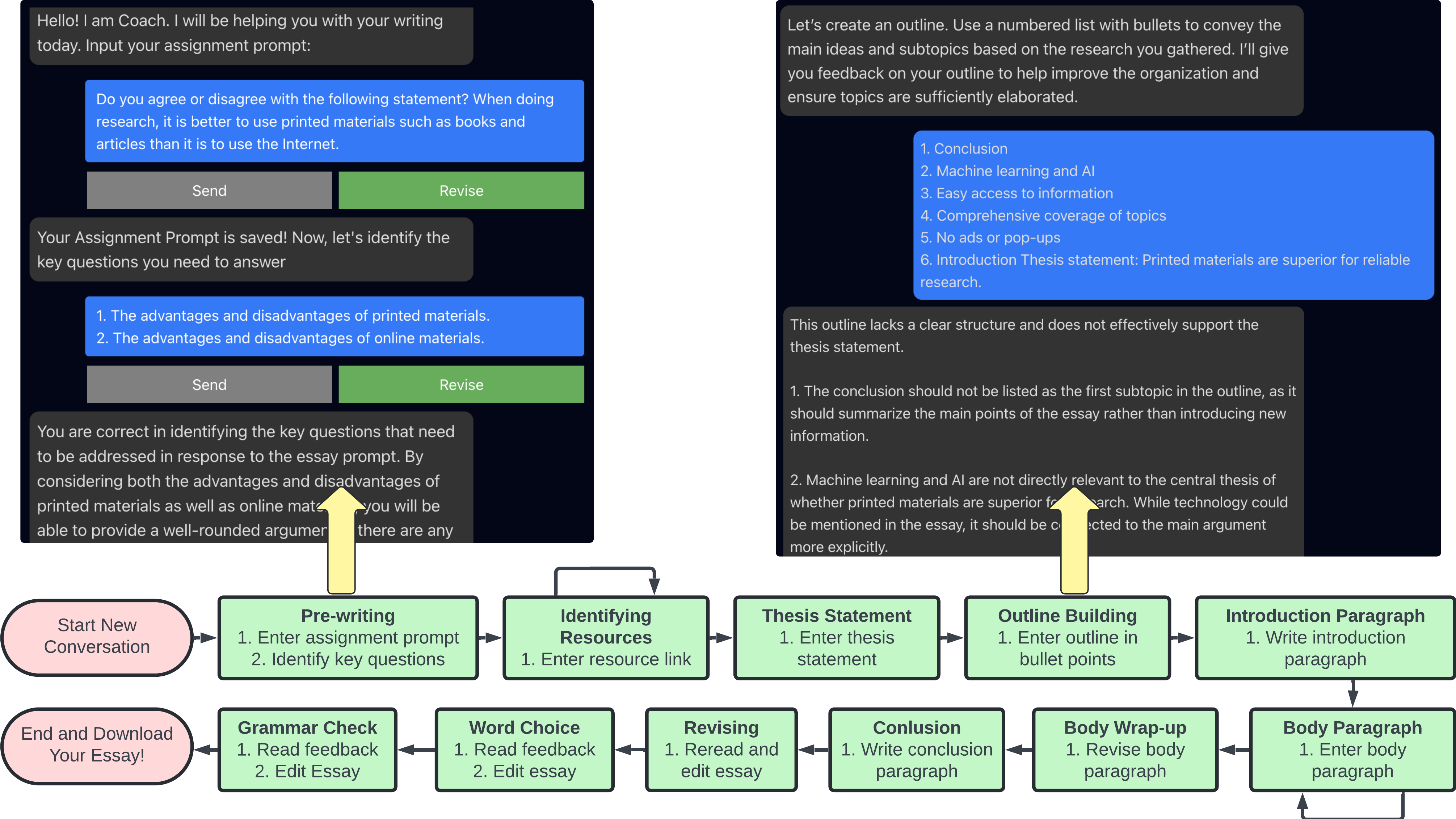}
    \caption{CoachGPT's Scaffolding Structure}
    \label{fig:scaffolding-structure}
    \vspace{-2mm}
\end{figure*}

\section{Introduction}
Academic writing is one of the essential tasks for students. It can be dispiriting without proper instructions and guidance, especially when writing in a second language \cite{silva1993toward}. Writers might get stuck on generating ideas, miss directions and planning, or make grammar mistakes. Traditionally, novice writers have few options to seek help on these roadblocks, such as asking writing instructors or searching dictionaries, which are not universally accessible. Therefore, writing assistant tools have been invented to help more writers. Early writing assistants are rule-based, focusing on detecting misspellings, subject-verb disagreements, and basic punctuation errors. Even though they can operate offline, they provide a low user experience with inaccurate, high false positives, and no contextual understanding. Various machine learning-based systems and products have been introduced with the advancement of natural language processing (NLP) and machine learning. The new systems and products are more intelligent in understanding context, tone, and writing style, but are more expensive and fail to understand domain-specific language. For example, tools like Grammarly require a subscription \footnote{\url{https://app.grammarly.com/}}.

Recently, as large language models (LLMs) dominate the mainstream natural language understanding tasks, writing assistants have become more intelligent \cite{brown2020language}. ChatGPT and DeepSeek are popular LLM-based AI platforms that answer general questions, and many writers use these platforms to assist their writing. However, existing LLMs have a fundamental limitation in education: they generate essays without teaching. Furthermore, LLMs can have detrimental effects on learning when misused. They are typically not trained with education-focused data, and their implementations are not geared toward education needs. Specifically, LLMs can encourage plagiarism and undermine authentic learning. Many news reports about students using ChatGPT for their essay writing and assignments have raised concerns about ethical issues. The ability to quickly generate answers to questions can create an undesirable learning shortcut, potentially diminishing students’ motivation to invest effort in their learning endeavors.

To address the limitations,  we developed {\em CoachGPT}, an LLM-based academic writing assistant that utilizes a scaffolding strategy ~\cite{Nordlof:14,Vygotsky:78}  to help writers succeed, from generating ideas and outline building to grammar and word choice checks. Besides, many writers feel distracted when using AI tools since AI responses are plain and lack human-like interactions. Hence, we incorporate personalized messages to make writers feel immersed. Our user study shows the effectiveness of CoachGPT in providing responsible writing advice through the scaffolding structure. Our user-friendly interface makes CoachGPT easy to use, and CoachGPT's personalized output makes our users feel attentive and adaptive. 

CoachGPT is accessible at \url{https://infochain.ece.udel.edu/coachgpt/}.

\vspace{-2mm}

\section{Related Work}
Early writing assistant systems employed rule‐based methods to address simple tasks such as spelling, punctuation, and basic grammatical error checks \cite{flower1981cognitive}, which are inaccurate and lack context awareness. Academic writing assistants have recently evolved significantly with the advancement of natural language processing and understanding. 

Grammarly and Educational Testing Service (ETS) were among the first to combine rule‑based corrections with machine‑learning models to evaluate and suggest grammar, tone, and style. However, their comprehensive error detection and usability are designed for evaluation purposes \cite{fitria2021grammarly}, which limits their ability to elevate writers' writing skills. That is, refreshing on grammar mistakes, tone, and style improvements does not necessarily improve writing skills. Moreover, Grammarly's subscription‑based pricing can be a barrier for some users. Other groups of ML-based assistants focus on text completion to help writers. Write-righter \cite{liu2016write} uses ML-based sentence extension to provide real-time hints and recommendations by analyzing the input context. OKI \cite{resch2019open} uses ML to retrieve relevant information to assist the initial conceptual stage of writing.

Recent progress in LLMs has spurred the development of more sophisticated academic writing assistants, and various tools have emerged. Writers might directly seek help from LLM models or platforms, for example, by asking ChatGPT about writing suggestions and grammar checking. However, LLMs are known to have hallucination issues and to forget more extended context. Enhancements are needed to make LLMs more helpful for writing \cite{mahmoudi2020rubrics}. Jasper AI \footnote{\url{https://www.jasper.ai}} is a writing assistant that leverages LLMs to help users write blog posts and market advertisements. Even though it is one of the best-known AI writing tools in the industry, it is a commercial tool that focuses on business scenarios and is unsuitable for academic writing, let alone its price. On the other hand, Jenni AI \footnote{\url{https://jenni.ai}} is an AI‑powered academic writing assistant designed to support researchers and students in streamlining the writing processes, especially helpful in managing citations and summarizing them. However, Jenni AI is designed to help advanced writers who already know how to write an academic paper. Therefore, it does not provide detailed instructions for novice writers. Although it provides feedback on user prompts, it offers minimal assistance to new writers.

More recent work has begun to address explainability, focused feedback, and discipline-specific fine-tuning in AI writing assistants. These include explainable interfaces that rationales \cite{kim2023towards}, a multi‑LLM pipeline that generates concise, actionable review comments to pinpoint weaknesses in scientific manuscripts \cite{chamoun2024automated}, fine‑tune LMs to support scientific authors through sentence scoring, section classification, and context‑aware paraphrasing \cite{mucke2023fine}, and domain‑specific LLM adaptations for in‑domain tasks \cite{cao2024scholargpt}. 

Previous studies of using AI as academic writing assisting tools can be categorized into three categories: as a brainstorming tool, as a text structuring tool, as a text polishing tool, and as an on-demand feedback tool \cite{sol2024ai}. However, none combine end‑to‑end scaffolding, real‑time feedback, and personalized interaction the way CoachGPT does. Therefore, we introduce CoachGPT, an LLM-powered academic writing assistant that provides scaffolding-like instructions to novice writers to assist in writing, from generating ideas to grammar and word choice checks, and also makes writers feel immersed by providing personalized interactions. 

\section{System Architecture}
CoachGPT aims to provide writers with an immersive, personal experience where they can feel more engaged and receive human-like feedback. Therefore, as shown in Figure \ref{fig:scaffolding-structure}, we designed a scaffolding structure to guide writers in academic writing, from generating ideas, identifying resources, and building outlines to paragraph writing and grammar checks.

\subsection{The scaffolding structure}

Scaffolding refers to a process where instruction is designed to meet learners at their current ability level and gradually build up their understanding and skills to reach a desired learning goal \cite{Lin:12}. In educational settings, scaffolding can manifest through various means, such as guided instruction, providing feedback, modeling a task, or gradually reducing the level of support as learners become more competent, with the goal of students independently executing the task in the future \cite{Belland:17}.

This section will explain in detail how the scaffolding structure  ~\cite{Nordlof:14,Vygotsky:78} is designed, which consists of 11 stages, as shown in Figure \ref{fig:scaffolding-structure}, including \textit{Pre-writing}, \textit{Identifying Resources}, \textit{Thesis Statement}, \textit{Outline Building}, \textit{Introduction Paragraph}, \textit{Body Paragraph}, \textit{Body Paragraph Wrap-up}, \textit{Conclusion Paragraph}, \textit{General Revising}, \textit{Word Choice Evaluation}, and finally \textit{Grammar Check}. Users must follow a linear order for these stages. CoachGPT fine-tuned messages at each stage based on our user study to ensure a personal and immersive writing experience.

Once users log in, they will be prompted to start a new writing task. We first ask users to input their assignment prompt, and CoachGPT saves their writing goal. The next step is identifying key questions based on the assignment prompt. CoachGPT will ask the user to input their key questions and then provide feedback by comparing the alignment between the assignment prompt and user input. Users can revise according to the feedback as many times as they want until they are satisfied with the key questions they would like to pursue. Then, CoachGPT will guide users to the following step: identifying resources, which are essential for academic writing. In this stage, one of the challenges is that writers fail to identify reliable resources, making their writing lack evidence. Therefore, we asked our large language model to evaluate the reliability of resource links based on domains and URLs. We hope users spend a reasonable amount of time finding relevant information regarding their assignment. Therefore, the system will be on hold until the users click on the next step. After users obtain enough information to write, they will be prompted to enter their thesis statement or topic. Like in the previous stage, CoachGPT will provide detailed feedback, and users can revise their statements as needed. Then, we encourage users to form an outline of their essay, and CoachGPT will provide suggestions on improving it. 

Now writers are ready to start writing the essay with identified key questions, a thesis statement, and a constructive outline. We break down writing into three sections: introduction, body, and conclusion, which align with most essay structures. We have a very straightforward user interface that lets writers enter these paragraphs and provides feedback and improvement suggestions for each section. Writers can revise on the spot and receive updated feedback to ensure they are on the right track. Finally, they can wrap up all the writing and form a complete essay. 

The final components in the scaffolding structure are the word choice and grammar check stages. It works similarly to existing tools like Grammarly. Writers will see highlights of their essays regarding world choice improvement and grammar mistakes.

\vspace{-4mm}
\subsection{Prompt Engineering}
In designing effective scaffolding experiences, we employed various prompt engineering techniques:

\textbf{Persona Prompting:} This approach shaped the responses to align with the role of a writing coach, enhancing performance on high-openness tasks, such as brainstorming and advice-giving \cite{olea2024persona}. For example, prompts explicitly instructed the model to \textit{"Act as a writing coach"} and provided some characteristics of a writing coach. This role-based prompting was intended to foster a coaching dynamic that supports learning without overriding student autonomy.

\textbf{Limiters/Constraints:} Incorporating limiters in prompts, such as explicitly instructing the AI not to rewrite students' work, ensured that feedback remained within educational boundaries \cite{schulhoff2024prompt}. For instance, one prompt stated: \textit{"You must not suggest any ideas or examples for the essay,"} reinforcing the AI's role as a guide rather than a content generator.

\textbf{Criteria-Based Feedback:} We combined criteria-based feedback to provide targeted guidance \cite{mahmoudi2020rubrics}. For example, prompts instructed the AI to assess a thesis statement based on criteria such as \textit{"off-topic, logical, and strong"}. This allows students to self-assess their work and address patterns of error.

\textbf{Output Presentation:} We designed the output structure to reduce cognitive load and enhance clarity \cite{sweller2011cognitive}. Prompts instructed the AI to break down feedback into sections based on different criteria (e.g., coherence, cohesion, clarity) and use simple language. One prompt stated: \textit{"Provide your response on the criteria in this order: spelling, grammar, and punctuation"}.

\textbf{Input Validation:} Input validation ensures that students engage meaningfully with the tool. A prompt included: \textit{"If the user does not type any paragraph or just random text, please direct them to type the paragraph."} This technique helps maintain the integrity of student submissions and ensures constructive interaction with the tool.
\vspace{-2mm}
\section{Implementation}

\begin{figure}
    \centering
    \includegraphics[width=1\linewidth]{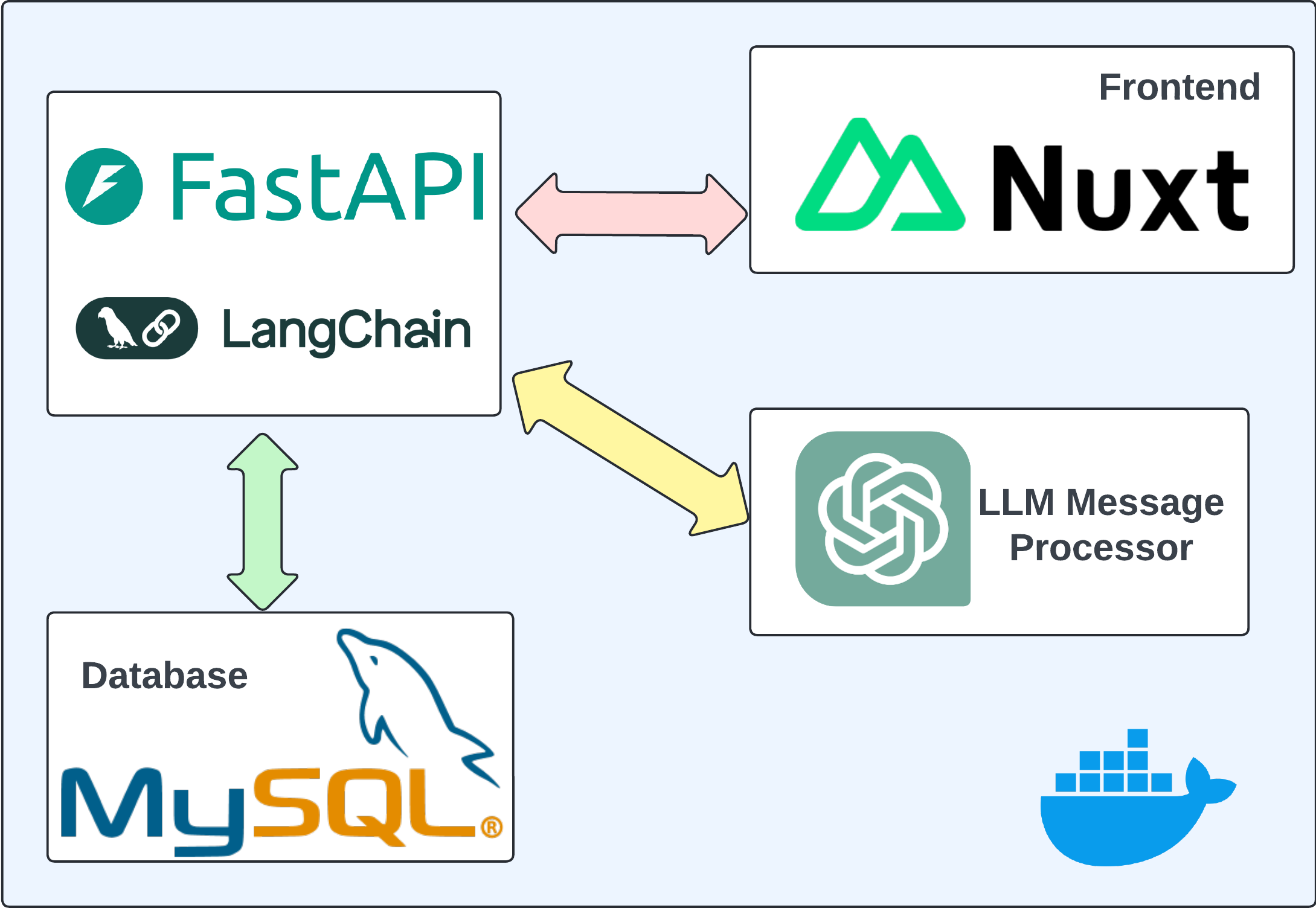}
    \caption{CoachGPT's Tech Stack}
    \label{fig:tech-stack}
\end{figure}

CoachGPT is a web-based application that can be run on any device with an Internet connection. Its implementation consists of three major parts: a front-end interface, a back-end processor, and a database; we packaged the three parts in \textit{Docker} and secured their connection as shown in Figure \ref{fig:tech-stack}. 

\vspace{-2mm}
\subsection{Frontend}
The frontend lets users interact with CoachGPT, including chatting with LLMs, asking questions, receiving feedback, and loading files. Several tools and templates can be used to build Chatbot applications. Given our unique design of the scaffolding structure, we decided to develop the interface from scratch. Front-end interface is created with an intuitive
\textit{Vue.js} \footnote{\url{https://vuejs.org/}} framework, \textit{Nuxt} \footnote{\url{https://nuxt.com/}}. We embedded a tutorial video on the home page to demonstrate our system's use.

\vspace{-2mm}
\subsection{Backend}
The backend is CoachGPT's brain. It processes user input and generates responses accordingly. The backend also handles communication between the front end and the database. All these communications are achieved with the backend APIs such as \textit{FastAPI} \footnote{\url{https://fastapi.tiangolo.com/}} and \textit{LangChain} \footnote{\url{https://www.langchain.com/}}. 

An AI assistant expects multiple rounds of user-system interactions, which are challenging for many language models since they tend to forget. Therefore, for memory persistence, we use \textit{LangChain} to store short-term memory and a database to store long-term memory. \textit{LangChain} also handles the modularity so that at each scaffolding stage, CoachGPT generates in-context responses and takes corresponding actions.  

\vspace{-2mm}
\subsection{Database}
A \textit{MySQL} database stores chat history (long-term memory) and user information. We chose MySQL as our database since all of the data is structured. Three tables are created: users, messages, and conversations, to store user information, message history, and conversation history. User interactions with the system are stored for user studies. Database operations are achieved by the Create, Read, Update, and Delete (CRUD) API. When LLM needs to reference long-term history, a query will be executed to retrieve relevant information from the database.

\vspace{-2mm}
\section{User Study}
 Cognitive interviews were conducted with Alex and Taylor (pseudonyms), with some experience in academic writing, and Jamie, with minimal experience. Content analysis was used to qualitatively analyze user experience, suitable for studying phenomena or cases with limited existing knowledge or literature \cite{hsieh2005qualitative}.

\textbf{Personalized output:} Users expressed that CoachGPT provided a sense of being heard and understood during their writing process. They appreciated how the platform responded to their input with personalized feedback, acknowledging their revisions and offering suggestions tailored to their needs and content. Taylor noted that the responses felt attentive and adaptive, adjusting to changes made within the text. 

\textbf{Navigation and Interface:} Users found the navigation intuitive but suggested improvements, such as viewing the outline during writing with an option to hide or unhide it. 

\textbf{Scaffolding Needs:} Jamie struggled with the term “key questions” in the pre-writing phase, requesting simpler approaches. All users appreciated more formulaic templates for thesis-building. Alex noted issues with counterargument detection. Moreover, the users believed the grammar feedback provided valuable insights but lacked specific error locations or in-text highlights offered by external tools, such as Grammarly.

\vspace{-2mm}
\section{Conclusion and Future Work}

In this work, we built CoachGPT, an AI-powered academic writing assistant that innovatively guides writers with scaffolding instructions and provides an immersive writing experience with personalized interactions. Compared with existing writing assistant tools, CoachGPT is explicitly designed to assist academic English writers with idea generation, outline building, and writing error detection. Our user study proves the effectiveness of CoachGPT in assisting academic writing with its user-friendly interface and attentive writing experience.

For future work, we will enhance the functionality and usability of the system based on users' feedback.  It would also be interesting to study how to generate writing prompts based on different writing domains and writing levels. 


\vspace{-2mm}
\begin{acks}
This work is supported by the IFSA and the AI Center of Excellence at the University of Delaware. We also thank the reviewers for their invaluable comments and suggestions. 
\end{acks}

\bibliographystyle{ACM-Reference-Format}
\bibliography{sample-base}


\end{document}